\documentclass{article} 
\usepackage{iclr2027_conference,times}

\usepackage{amsmath,amsfonts,bm}

\def\eqref#1{equation~\ref{#1}}

\def\1{\bm{1}}

\DeclareMathAlphabet{\mathsfit}{\encodingdefault}{\sfdefault}{m}{sl}
\SetMathAlphabet{\mathsfit}{bold}{\encodingdefault}{\sfdefault}{bx}{n}

\usepackage{hyperref}
\usepackage{url}
\usepackage{graphicx}
\usepackage{algorithm}
\usepackage{algorithmic}
\usepackage{amssymb}
\usepackage{multirow}
\usepackage{amsmath}
\usepackage{subcaption}
\usepackage{booktabs}
\iclrfinaltrue

\title{Experts Exhibit Structured Redundancy: Structure-Preserving Expert Pruning for Sparse Mixture-of-Experts Large Language Models}

\author{
Tian Wang\textsuperscript{1,2}\quad
Yongding Tao\textsuperscript{1,2}\quad
Yihong Dong\textsuperscript{3}\quad
Xiaoming Wu\textsuperscript{4}\quad
Ying Guo\textsuperscript{5}\quad
Ge Li\textsuperscript{1,2}
\\[0.7em]
\textsuperscript{1}Peking University
\\[-0.1em]
\small \textsuperscript{2}Key Laboratory of High Confidence Software Technologies,
Peking University, Ministry of Education
\\[-0.1em]
\textsuperscript{3}Shanghai Jiaotong University
\\[-0.1em] 
\textsuperscript{4}Qilu University of Technology
\\[-0.1em] 
\textsuperscript{5}Shandong Computer Science Center
}

\begin{document}

\maketitle

\begin{abstract}
Sparse Mixture-of-Experts (MoE) large language models replace the dense feed-forward networks in standard Transformers with MoE layers consisting of multiple experts, activating only a small subset of experts for each token during the forward pass. This sparse activation mechanism significantly improves the trade-off between model capacity and inference efficiency. However, practical deployment still requires storing all experts, resulting in substantial memory overhead. In this work, we show that expert outputs in MoE layers exhibit structured redundancy. Across different models and task categories, the stable rank and entropy effective rank of expert outputs are consistently far below the nominal number of experts, while expert output distributions also exhibit fewer effective modes. Motivated by these observations, we reformulate expert pruning as a structure-preserving expert subset selection problem and efficiently optimize it via a greedy strategy, rather than only independently scoring and ranking individual experts or exhaustively searching over expert combinations. We propose two expert pruning methods under the unified structure-preserving framework: Projection Energy Expert Pruning (\textbf{PEEP}), which preserves the functional subspace by maximizing weighted projection energy, and Distributional Coverage Expert Pruning (\textbf{DCEP}), which preserves the distributional geometry by minimizing weighted coverage error. Both methods incorporate router-derived weights as priors, enabling them to jointly preserve the global structural relationships among experts while accounting for the local importance of individual experts. Experiments on DeepSeek-V2-Lite and Qwen3-30B-A3B demonstrate that, at the expert retention ratios of $r{=}75\%$ and $r{=}50\%$, our methods outperform the state-of-the-art baselines on the majority of benchmarks and consistently achieve the best average performance across both general tasks and mathematical reasoning tasks.
\end{abstract}

\section{Introduction}
Mixture-of-Experts (MoE) has become an important architectural paradigm for scaling large language models (LLMs)~\citep{LLM}. Sparse MoE models replace the dense feed-forward networks (FFNs) in standard Transformers~\citep{Transformer} with multiple experts and route each token to only a small subset of experts in each layer. This design enables scaling up the total number of model parameters and improving model capacity while keeping the number of activated parameters during inference. Despite their computational sparsity, MoE LLMs remain expensive to deploy, since all expert parameters must be stored during inference. As experts dominate the number of parameters, expert-level pruning becomes an increasingly important direction for MoE compression. 

Existing expert pruning methods usually score and rank experts according to their individual importance based on activation frequency, router scores, activation magnitudes, or other hand-crafted metrics, and then remove the less important experts~\citep{TSEP,EEP,MoE_Pruner,REAP}. However, these approaches overlook that experts do not contribute to the model independently. The quality of a retained expert subset depends not only on the individual importance of each expert, but also on whether the selected experts jointly preserve the structural characteristics of the original MoE layer. Several recent studies have attempted to consider inter-expert interactions in expert pruning~\citep{NAEE,MoE_I2,EEP}. However, they typically identify expert subsets through exhaustive enumeration or heuristic combinatorial optimization, introducing substantial pruning overhead. Such approaches are difficult to scale to modern MoE LLMs with tens or hundreds of experts per layer.

In this work, we reformulate expert pruning as a structure-preserving expert subset selection problem. Our empirical analysis in Figure~\ref{fig:motivation} reveals that expert outputs in MoE layers exhibit strong structured redundancy. From a functional subspace perspective, both the stable rank and entropy effective rank of expert output representations are substantially lower than the number of experts. From a distributional perspective, expert output distributions also contain fewer effective modes than the number of experts. These observations motivate a new pruning framework that selects a compact yet complementary expert subset to preserve the structural characteristics of the original MoE layer. 

We instantiate this framework with two expert pruning methods that avoid exhaustive enumeration and costly heuristic search. Projection Energy Expert Pruning (PEEP) views expert outputs as basis functions and greedily selects experts that maximize weighted projection energy, thereby preserving the dominant functional subspace structure. Distributional Coverage Expert Pruning (DCEP) represents expert outputs as Gaussian distributions and greedily selects medoid experts that minimize weighted coverage error, thereby preserving distributional geometry. Both methods incorporate router-derived weights as priors, enabling them to jointly preserve global structural relationships among experts and local importance of individual experts.

Our contributions are summarized as follows:
\begin{itemize}
\item We reveal structured redundancy in expert outputs from both functional subspace and distributional geometry perspectives, showing that the effective structure of expert representations is substantially smaller than the total number of experts.
\item Unlike prior pruning methods that rank experts individually, we reformulate expert pruning as a structure-preserving expert subset selection problem and efficiently solve it without resorting to expensive combinatorial search over expert subsets.
\item We instantiate the structure-preserving framework with two expert pruning methods, PEEP and DCEP. The former preserves the functional subspace structure, while the latter preserves the distributional geometry. Both methods are implemented through efficient greedy selection. 
\item Experiments on DeepSeek-V2-Lite and Qwen3-30B-A3B demonstrate that our methods outperform the state-of-the-art (SOTA) baselines on the majority of benchmarks and consistently achieve the best average performance across both general tasks and mathematical reasoning tasks.
\end{itemize}

\section{Related Work}
\paragraph{MoE Architectures}
MoE architectures scale model capacity while routing each token to a small subset of experts. Early work introduced learned sparse routing for conditional computation~\citep{Sparse_MoE}, followed by GShard~\citep{GShard} and Switch Transformer~\citep{Swtich_Transformers}, which improved large-scale training. Mixtral~\citep{Mixtral} showed that sparse MoE can deliver competitive performance with markedly lower inference cost. Recent MoE LLMs further enhance expert specialization through shared experts in DeepSeekMoE~\citep{DeepSeekMoE} and fine-grained experts in Qwen3~\citep{Qwen3}.

\paragraph{MoE Compression}
MoE compression reduces the number of parameter and memory overhead introduced by storing all experts. Expert pruning directly removes complete experts. Many existing methods score and rank experts according to their individual importance based on routing or activation statistics. TSEP performs task-specific progressive pruning during fine-tuning~\citep{TSEP}. Subsequently, SEER-MoE combines routing statistics with regularized fine-tuning~\citep{SEER_MoE}, while MC-Suite systematically investigates a broad range of expert importance metrics~\citep{jaiswal2025fantastic}. EASY-EP estimates domain-specific expert importance using router scores and similarity~\citep{EASY_EP} of expert output, whereas REAP combines router scores with activation norms~\citep{REAP}. Although efficient, these methods primarily rank experts independently. Other expert pruning methods consider inter-expert interactions more explicitly. NAEE selects expert combinations by minimizing layer-output reconstruction error through subset enumeration~\citep{NAEE}. EEP and MoE-I$^2$ employ evolutionary or genetic search to optimize the retained expert subsets~\citep{EEP,MoE_I2}. Shapley-MoE~\citep{Shapley_MoE} estimates each expert's contribution using Shapley values and removes experts with relatively small contributions. These methods capture inter-expert interactions but may incur substantial search costs. Beyond pruning, STUN clusters experts according to behavioral similarity and then applies expert merging and unstructured pruning~\citep{STUN}. MC-SMoE aligns and merges experts using routing-derived clusters~\citep{MC}, HC-SMoE performs expert merging through hierarchical clustering~\citep{HC_SMoE}, ConDense-MoE condenses sparse MoE layers into compact dense layers~\citep{Condense}, and MoNE replaces redundant experts with lightweight novice modules~\citep{MoNE}. Unlike methods that score and rank experts individually, search over possible expert subsets, or merge experts by modifying their parameters, our methods efficiently select and retain a subset of the original experts.

\section{Preliminaries}
\paragraph{MoE Layer} A standard Transformer layer contains a self-attention module followed by a FFN. In a sparse MoE Transformer, the dense FFN is replaced by $N$ expert FFNs, denoted as ${E_i}$. Given the hidden state $\mathbf{h}_x \in \mathbb{R}^{d}$ of a token $x$, a linear router $\mathbf{W}_{\mathrm{router}}$ computes the logit for each expert:
\begin{equation}
s_i = [\mathbf{W}_{\mathrm{router}}\mathbf{h}_{x}]_i, \mathbf{W}_{\mathrm{router}} \in \mathbb{R}^{N \times d}.
\end{equation}
To reduce computation, the MoE layer activates only the experts with the highest top-$k$ logits and normalizes them into routing scores
\begin{equation}
\alpha_i = \frac{\exp(s_i)}{\sum_{j\in\mathcal{T}_x}\exp(s_j)},
\end{equation}
where $\mathcal{T}_x$ is the selected expert indices. The output of the MoE layer is then computed as a weighted sum of the outputs of the selected experts
\begin{equation}
\mathbf{h}^{'}_x=\sum_{i\in\mathcal{T}_x}\alpha_iE_i(x),
\end{equation}

\paragraph{Expert Pruning}
Expert pruning aims to permanently remove a subset of experts from each MoE layer. Formally, for layer $\ell$ with $N$ experts and a target retention ratio $r$, the goal is to select a retained expert subset
\begin{equation}
\label{eq:expert_retention}
S^{(\ell)}\subseteq\{1,\ldots,N\},
\quad|S^{(\ell)}|=K=\lfloor rN\rfloor,
\end{equation}
and remove the other experts.

\begin{figure}[t]
\centering
\includegraphics[width=0.9\textwidth]{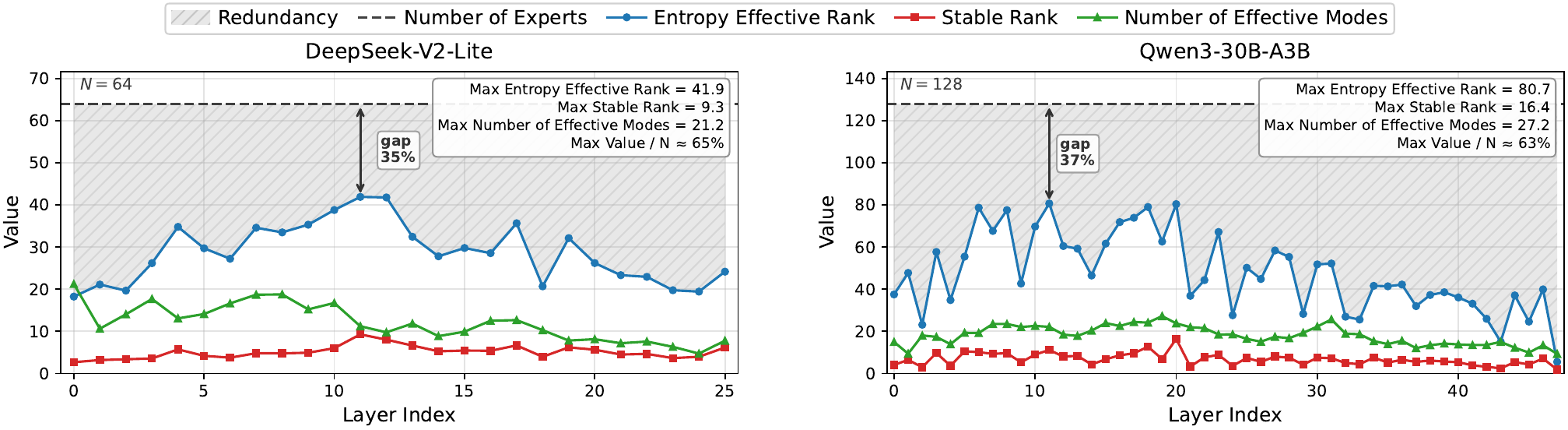}
\caption{
Ranks of expert representations and numbers of effective modes of expert output distributions across MoE layers.
\textit{Left:} DeepSeek-V2-Lite with 64 experts per layer.
\textit{Right:} Qwen3-30B-A3B with 128 experts per layer.
Across all MoE layers, the entropy effective rank, stable rank, and number of effective modes remain well below the nominal number of experts, reaching at most 63\%--65\% of the total number of experts and thereby revealing substantial structured redundancy in expert outputs.
}
\label{fig:motivation}
\end{figure}

\section{Motivation}
\label{sec:motivation}
The output of an MoE layer is produced by a routed aggregation of multiple expert outputs, suggesting that expert redundancy can be understood in terms of the global structure formed by the outputs of all experts. We therefore investigate whether the expert outputs exhibit structural redundancy from two perspectives: \emph{functional subspace} and \emph{distributional geometry}.

\paragraph{Calibration Statistics}
Given a calibration dataset $\mathcal{D}$, we perform a forward pass through the model and collect the output of all experts. For the $i$-th expert in MoE layer $\ell$, let $f_i^{(\ell)}(\mathbf{h}_x)$ denote the output for the hidden state $\mathbf{h}_x$ of token $x$. We estimate the mean of each expert's output as
\begin{equation}
\boldsymbol{\mu}_i^{(\ell)}
=
\mathbb{E}_{x \sim \mathcal{D}}
\left[
f_i^{(\ell)}(\mathbf{h}_x)
\right].
\end{equation}
We additionally estimate the standard deviation of each expert's output as
\begin{equation}
\boldsymbol{\sigma}_i^{(\ell)}
=
\sqrt{
\mathbb{E}\left[\mathbf{z}_i^2\right]
-
\left(\mathbb{E}\left[\mathbf{z}_i\right]\right)^2
}, \mathbf{z}_i=f_i^{(\ell)}(\mathbf{h}_x).
\end{equation}

\paragraph{Low-Dimensional Functional Subspace}
We use the mean of each expert's output as its representation and investigate whether the expert output representations are concentrated in a lower-dimensional subspace. For each MoE layer $\ell$ with $N$ experts, we construct the representation matrix and compute its singular value decomposition
\begin{equation}
\label{eq:expert_representation_matrix}
\mathbf{W}^{(\ell)}
=
\begin{bmatrix}
(\boldsymbol{\mu}_1^{(\ell)})^\top
&
\cdots
&
(\boldsymbol{\mu}_N^{(\ell)})^\top
\end{bmatrix}^{\!\top}
=
\mathbf{U}\boldsymbol{\Sigma}\mathbf{V}^{\top},
\mathbf{W}^{(\ell)}
\in \mathbb{R}^{N \times d}.
\end{equation}
We get the singular values $\sigma_1 \geq \sigma_2 \geq \cdots$. Then we quantify the effective dimensionality of expert output representations using the entropy effective rank based on normalized spectral energy
\begin{equation}
r_{\mathrm{entropy}}
=
\exp\left(
-\sum_j p_j \log p_j
\right), p_j
=
\frac{\sigma_j^2}
{\sum_k \sigma_k^2},
\end{equation}
and the stable rank
\begin{equation}
r_{\mathrm{stable}}
=
\frac{\|\mathbf{W}^{(\ell)}\|_F^2}
{\|\mathbf{W}^{(\ell)}\|_2^2}
=
\frac{\sum_j \sigma_j^2}{\sigma_1^2}.
\end{equation}
Both metrics characterize the effective dimensionality of the subspace spanned by the expert output representations and are less sensitive than the algebraic rank to small singular values arising from numerical noise.

As shown in Figure~\ref{fig:motivation}, both metrics are substantially lower than the number of experts across all MoE layers. These results indicate that the expert output representations are 
in a lower-dimensional subspace, suggesting substantial redundancy in the function space spanned by the expert outputs. Therefore, it may be unnecessary to preserve all experts to maintain the function space. Instead, a compact subset of complementary experts may suffice to preserve the original function space.

\paragraph{Redundant Distributional Geometry}
The mean of each expert's output does not capture the distributional variation. We therefore examine the distributional geometry formed by the expert outputs and investigate whether different experts have similar distributional modes. We approximate the output distribution of expert $i$ by a diagonal Gaussian
\begin{equation}
\label{eq:expert_distribution}
P_i^{(\ell)}
=
\mathcal{N}
\left(
\boldsymbol{\mu}_i^{(\ell)},
\operatorname{diag}
\left(
(\boldsymbol{\sigma}_i^{(\ell)})^2
\right)
\right).
\end{equation}
For two experts $i$ and $j$, we measure their distributional discrepancy using the squared 2-Wasserstein distance
\begin{equation}
\label{eq:expert_distance}
d_{i,j}^{(\ell)}
=
\Vert
\boldsymbol{\mu}_i^{(\ell)}
-
\boldsymbol{\mu}_j^{(\ell)}
\Vert_2^2
+
\Vert
\boldsymbol{\sigma}_i^{(\ell)}
-
\boldsymbol{\sigma}_j^{(\ell)}
\Vert_2^2.
\end{equation}
We convert pairwise distances into a similarity kernel
\begin{equation}
K_{i,j}^{(\ell)}
=
\exp
\left(
-\frac{d_{i,j}^{(\ell)}}{c\cdot\tau^{(\ell)}}
\right),
\end{equation}
where $\tau^{(\ell)}$ is the median of the pairwise distances within layer and $c$ is the scaling factor. The numbers of effective modes under different values of $c$ are provided in Figure~\ref{fig:ab_neff} in the appendix.

To estimate the effective number of distinct distributional modes, let $\lambda_1,\ldots,\lambda_N$ denote the eigenvalues of $\mathbf{K}^{(\ell)}$. We compute their participation ratio as
\begin{equation}
N_{\mathrm{eff}}^{(\ell)}
=
\frac{
\left(\sum_{j=1}^{N}\lambda_j\right)^2
}{
\sum_{j=1}^{N}\lambda_j^2
}.
\end{equation}
If expert distributions are mutually distinct, the similarity kernel approaches the identity matrix and $N_{\mathrm{eff}^{(\ell)}}$ approaches $N$. In contrast, strongly similar expert distributions induce a spectrally concentrated kernel and yield $N_{\mathrm{eff}}^{(\ell)} \ll N$. As shown in Figure~\ref{fig:motivation}, the effective number of distributional modes is consistently much smaller than the total number of experts. These results indicate that expert distributions contain a limited number of representative modes, with multiple experts providing overlapping coverage. A compact subset of experts is sufficient to preserve most of the coverage of the original expert distribution.

\begin{figure*}[t]
\centering
\includegraphics[width=0.98\textwidth]{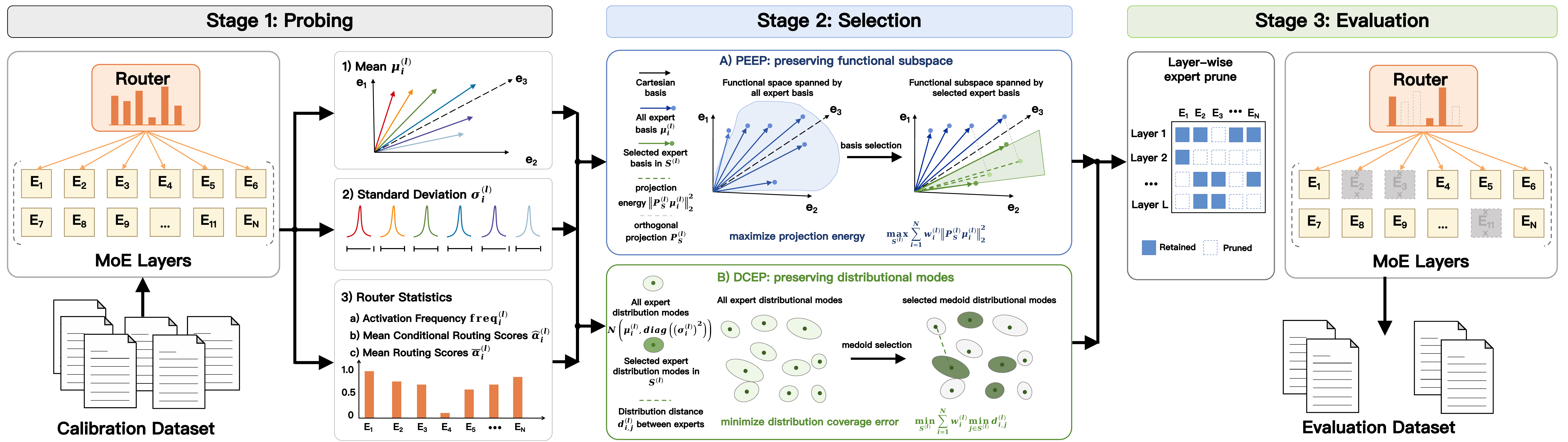}
\caption{Overview of structure-preserving MoE expert pruning.
\textit{Probing:} collect expert outputs and router statistics on calibration dataset.
\textit{Selection:} PEEP maximizes projection energy; DCEP minimizes distributional coverage error.
\textit{Evaluation:} prune experts, restrict routing to the retained expert set and evaluate the pruned MoE LLMs on downstream tasks.}
\label{fig:method}
\end{figure*}

\vspace{-0.5em}
\section{Method}
The observations in the previous section reveal two forms of structured redundancy in MoE layers. First, the expert output representations are concentrated in a low-dimensional functional subspace, indicating that a subset of complementary experts can preserve the original function space. Second, expert output distributions exhibit a limited number of effective modes, suggesting that a subset of complementary experts is sufficient to capture the original distributional geometry. These findings motivate us to reformulate expert pruning as a \emph{structure-preserving subset selection} problem, where the goal is to identify a compact expert subset that maximally preserves the structure of the original MoE layer outputs. We instantiate this structure-preserving framework with PEEP and DCEP. Both methods select expert subsets that minimize structural deviation from the original expert set. Both methods use a greedy strategy, thereby enabling effective expert pruning. An overview of our methods is shown in Figure~\ref{fig:method}.

\subsection{Projection Energy Expert Pruning}
\paragraph{Optimization Objective}
For layer $\ell$, let $w^{(\ell)}_i \in \mathbb{R}^{+}$ denote router-derived weights for expert $i$ in layer $\ell$, we can define the weighted representation Gram matrix
\begin{equation}
    \mathbf{C}^{(\ell)} = \big(\mathbf{W}^{(\ell)}\big)^\top \,\mathrm{diag}\!\big(\boldsymbol{w}^{(\ell)}\big)\, \big(\mathbf{W}^{(\ell)}\big),
\end{equation}
where $\mathbf{W}^{(\ell)}$ is the expert representation matrix in Equation~\eqref{eq:expert_representation_matrix}, $\boldsymbol{w}^{(\ell)}=[w^{(\ell)}_1,...,w^{\ell}_N]^\top \in \mathbb{R}_{+}^{N}$.
For a retained expert set $S^{(\ell)} \subseteq \{1,\ldots,N\}$, let $\mathbf{P}_S^{(\ell)}$ be the orthogonal projector onto $\mathcal{W}_{S}^{(\ell)} = \mathrm{span}\{\boldsymbol{\mu}_i^{(\ell)} : i \in S^{(\ell)}\}$.
PEEP selects experts by minimizing the weighted projection error
\begin{equation}
    \mathcal{L}_{\mathrm{PEEP}}(S^{(\ell)})
    = \sum_{i=1}^{N} w_i^{(\ell)} \big\|\boldsymbol{\mu}_i^{(\ell)} - \mathbf{P}_S^{(\ell)} \boldsymbol{\mu}_i^{(\ell)}\big\|_2^2,
    \label{eq:peep-loss}
\end{equation}
which is equivalent to maximizing the weighted projection energy
$\sum_{i=1}^{N} w_i^{(\ell)} \big\|\mathbf{P}_S^{(\ell)} \boldsymbol{\mu}_i^{(\ell)}\big\|_2^2$.
Intuitively, PEEP seeks a subset of complementary basis experts whose span approximates the original function space.

\paragraph{Greedy Selection}
We adopt a greedy procedure that iteratively adds the expert yielding the largest marginal energy gain.
Let $\mathbf{Q} \in \mathbb{R}^{d \times t}$ be an orthonormal basis of $\mathcal{W}_S^{(\ell)}$ after $t$ selections.
For candidate $j \notin S^{(\ell)}$, define the residual direction
\begin{equation}
    \boldsymbol{q}_j \;\propto\; \big(\mathbf{I} - \mathbf{Q}\mathbf{Q}^\top\big) \boldsymbol{\mu}_j^{(\ell)},
    \qquad
    \|\boldsymbol{q}_j\|_2 = 1,
\end{equation}
which removes components of $\boldsymbol{\mu}_j^{(\ell)}$ already explained by $S^{(\ell)}$.
The marginal gain of adding expert $j$ is
\begin{equation}
    \Delta_{\mathrm{PEEP}}(j \mid S^{(\ell)})
    = \sum_{i=1}^{N} w_i^{(\ell)} \,\big(\boldsymbol{q}_j^\top \boldsymbol{\mu}_i^{(\ell)}\big)^2
    = \boldsymbol{q}_j^\top \mathbf{C}^{(\ell)} \boldsymbol{q}_j.
    \label{eq:peep-gain}
\end{equation}
At each step we choose $j^\star = \arg\max_{j \notin S^{(\ell)}} \Delta_{\mathrm{PEEP}}(j \mid S^{(\ell)})$, update $\mathbf{Q}\leftarrow[\mathbf{Q},\mathbf{q}_{j^\star}]$, and repeat until $|S^{(\ell)}|=K$, where $K$ is the number of retained experts in Equation~\eqref{eq:expert_retention}. Algorithm~\ref{alg:peep} summarizes the PEEP procedure.

\begin{algorithm}[t]
\small
\caption{Projection Energy Expert Pruning}
\label{alg:peep}
\textbf{Input}: expert representation matrix $\mathbf{W}^{(\ell)} \in \mathbb{R}^{N \times d}$, weights $\boldsymbol{w}^{(\ell)} \in \mathbb{R}^{N}$, number of retained experts $K$, tolerance $\epsilon$\\
\textbf{Output}: retained expert subset $S^{(\ell)}$
\begin{algorithmic}[1]
\STATE Normalize $\boldsymbol{w}^{(\ell)} \leftarrow \boldsymbol{w}^{(\ell)} / \mathrm{mean}(\boldsymbol{w}^{(\ell)})$
\STATE $\mathbf{C}^{(\ell)} \leftarrow \big(\mathbf{W}^{(\ell)}\big)^\top \,\mathrm{diag}(\boldsymbol{w}^{(\ell)})\, \mathbf{W}^{(\ell)}$
\STATE $S^{(\ell)} \leftarrow \emptyset$;\; $\mathbf{Q} \leftarrow [\ ] \in \mathbb{R}^{d \times 0}$
\FOR{$t = 1$ \TO $K$}
    \STATE $\Delta^\star \leftarrow -\infty$;\; $j^\star \leftarrow \texttt{null}$;\; $\boldsymbol{q}^\star \leftarrow \texttt{null}$
    \FOR{each $j \in \{1,\ldots,N\} \setminus S^{(\ell)}$}
        \STATE $\boldsymbol{v} \leftarrow \boldsymbol{\mu}^{(\ell)}_j$
        \STATE $\boldsymbol{v}_\perp \leftarrow \boldsymbol{v} - \mathbf{Q}\mathbf{Q}^\top \boldsymbol{v}$
        \IF{$\|\boldsymbol{v}_\perp\|_2 > \epsilon$}
            \STATE $\boldsymbol{q} \leftarrow \boldsymbol{v}_\perp / \|\boldsymbol{v}_\perp\|_2$
            \STATE $\Delta_j \leftarrow \boldsymbol{q}^\top \mathbf{C}^{(\ell)} \boldsymbol{q}$
            \IF{$\Delta_j > \Delta^\star$}
                \STATE $\Delta^\star \leftarrow \Delta_j$;\; $j^\star \leftarrow j$;\; $\boldsymbol{q}^\star \leftarrow \boldsymbol{q}$
            \ENDIF
        \ENDIF
    \ENDFOR
    \STATE $S^{(\ell)} \leftarrow S^{(\ell)} \cup \{j^\star\}$;\; $\mathbf{Q}\leftarrow[\mathbf{Q},\mathbf{q}^\star]$
\ENDFOR
\STATE \textbf{return} $S^{(\ell)}$
\end{algorithmic}
\end{algorithm}

\begin{algorithm}[t]
\small
\caption{Distributional Coverage Expert Pruning}
\label{alg:dcep}
\textbf{Input}: mean of each expert's output $\{\boldsymbol{\mu}_i^{(\ell)}\}$, std of each expert's output $\{\boldsymbol{\sigma}_i^{(\ell)}\}$, weights $\boldsymbol{w}^{(\ell)}$, number of retained experts $K$\\
\textbf{Output}: retained expert subset $S^{(\ell)}$
\begin{algorithmic}[1]
\STATE Z-score normalize $\boldsymbol{\mu}_i^{(\ell)}, \boldsymbol{\sigma}_i^{(\ell)}$ across experts for each dimension
\STATE Calculate $d^{(\ell)}_{i,j}$ via Eq.~\eqref{eq:expert_distance}
\STATE Normalize $\boldsymbol{w}^{(\ell)} \leftarrow \boldsymbol{w}^{(\ell)} / \mathrm{mean}(\boldsymbol{w}^{(\ell)})$
\STATE $S^{(\ell)} \leftarrow \emptyset$;\; $\delta_i \leftarrow +\infty$ for $i=1,\ldots,N$
\FOR{$t = 1$ \TO $K$}
    \STATE $\Delta^\star \leftarrow -\infty$;\; $j^\star \leftarrow \texttt{null}$
    \FOR{each $j \in \{1,\ldots,N\} \setminus S^{(\ell)}$}
        \IF{$t = 1$}
            \STATE $\Delta_j \leftarrow -\sum_{i=1}^{N} w_i^{(\ell)} d^{(\ell)}_{i,j}$
        \ELSE
            \STATE $\Delta_j \leftarrow \sum_{i=1}^{N} w_i^{(\ell)} \big[\delta_i - \min(\delta_i, d^{(\ell)}_{i,j})\big]$
        \ENDIF
        \IF{$\Delta_j > \Delta^\star$}
            \STATE $\Delta^\star \leftarrow \Delta_j$;\; $j^\star \leftarrow j$
        \ENDIF
    \ENDFOR
    \STATE $S^{(\ell)} \leftarrow S^{(\ell)} \cup \{j^\star\}$; $\delta_i \leftarrow \min(\delta_i, d^{(\ell)}_{i,j^\star})$ for $i=1,\ldots,N$
\ENDFOR
\STATE \textbf{return} $S^{(\ell)}$
\end{algorithmic}
\end{algorithm}

\subsection{DCEP: Distributional Coverage Expert Pruning}
\paragraph{Optimization Objective}
For layer $\ell$, let $w^{(\ell)}_i \in \mathbb{R}_{+}$ denote router-derived weights for expert $i$ in layer $\ell$. According to the distance matrix in Equation~\eqref{eq:expert_distance}, we define the coverage error of each expert $i$ by the distance to its nearest medoid in selected expert set $S^{(\ell)}$. DCEP selects experts by minimizing the weighted coverage error
\begin{equation}
    \mathcal{L}_{\mathrm{DCEP}}(S^{(\ell)})
    = \sum_{i=1}^{N} w_i^{(\ell)} \min_{j \in S^{(\ell)}} d^{(\ell)}_{i,j}.
    \label{eq:dcep-loss}
\end{equation}
Intuitively, DCEP seeks a subset of medoid experts whose distributional modes approximately cover the original distributional geometry.

\paragraph{Greedy Selection}
We also adopt a greedy procedure that iteratively adds the expert yielding the largest coverage gain. Let $\delta_i(S^{(\ell)}) = \min_{j \in S^{(\ell)}} d^{(\ell)}_{i,j}$ be the current nearest-medoid distance for expert $i$, with $\delta_i(\emptyset) = +\infty$.
When adding candidate $j \notin S^{(\ell)}$, the coverage gain is
\begin{equation}
    \Delta_{\mathrm{DCEP}}(j \mid S^{(\ell)})
    = \sum_{i=1}^{N} w_i^{(\ell)}
    \Big[\delta_i(S^{(\ell)}) - \min\!\big(\delta_i(S^{(\ell)}),\, d^{(\ell)}_{i,j}\big)\Big].
    \label{eq:dcep-gain}
\end{equation}
We select $j^\star = \arg\max_{j \notin S^{(\ell)}} \Delta_{\mathrm{DCEP}}(j \mid S^{(\ell)})$, update $\delta_i(S^{(\ell)}) \leftarrow \min(\delta_i(S^{(\ell)}), d^{(\ell)}_{i,j^\star})$, and repeat until $|S^{(\ell)}|=K$.
For $S^{(\ell)}=\emptyset$, $\Delta_{\mathrm{DCEP}}(j \mid \emptyset) = -\sum_i w_i^{(\ell)} d^{(\ell)}_{i,j}$. Algorithm~\ref{alg:dcep} summarizes the DCEP procedure.

\subsection{Router-Derived Weights}
Both PEEP and DCEP use the weights $w_i^{(\ell)}$ as a prior information of expert importance. We compute three router signals in the calibration forward pass:
\textbf{(i)}~\emph{activation frequency} $freq_i^{(\ell)}$, the fraction of tokens for which expert $i$ appears in $\mathcal{T}(x)$;
\textbf{(ii)}~\emph{mean conditional routing score} $\hat{g}_i^{(\ell)}$, the average router score of expert $i$ over the tokens for which expert $i$ appears in $\mathcal{T}(x)$;
\textbf{(iii)}~\emph{mean routing score} $\bar{g}_i^{(\ell)}$, the average router score of expert $i$ over all tokens.
Each signal is normalized to $(0,1]$ within the layer and multiplicatively combined, e.g.,
$w_i^{(\ell)} \propto freq_i^{(\ell)} \hat{g}_i^{(\ell)} \bar{g}_i^{(\ell)} $ for the full combination. Weights are also normalized to unit mean. Router-derived weights reflect the local importance of individual experts, complementing PEEP and DCEP which capture the global structure of expert outputs.

\section{Experiments}
\label{sec:experiments}
\subsection{Experimental Setup}
We evaluate our methods on DeepSeek-V2-Lite and Qwen3-30B-A3B at expert retention ratio of $r{=}75\%$ and $r{=}50\%$. We assess general capability on HellaSwag~\citep{HellaSwag}, ARC-Challenge~\citep{ARC}, MMLU~\citep{MMLU}, and C-Eval~\citep{CEval}, and mathematical reasoning capability on GSM8K~\citep{GSM8K}, MATH500~\citep{MATH500}, AIME24~\citep{AIME2024}, and AIME25~\citep{AIME2025}. Calibration uses 32 C4 samples~\citep{C4} for general tasks, 32 GSM8K training samples for GSM8K and MATH500, and the 30 AIME23 problems~\citep{AIME2023} for AIME24 and AIME25. The baselines include random pruning, frequency-based pruning, EASY-EP~\citep{EASY_EP} and REAP~\citep{REAP}. Detailed experimental information is provided in Appendix~\ref{app:experimental_setup}.

\begin{table*}[ht]
\centering
\caption{
Performance comparison of PEEP and DCEP with baseline expert pruning methods on DeepSeek-V2-Lite and Qwen3-30B-A3B across general and mathematical reasoning benchmarks at expert retention ratios of $r{=}75\%$ and $r{=}50\%$. The best and second-best results among pruned methods under each model and expert retention ratio are highlighted in bold and underlined, respectively. ``--'' indicates that the model fails to complete the corresponding task.
}
{
\resizebox{0.9\textwidth}{!}{
\begin{tabular}{llcccccccccc}
\specialrule{1.2pt}{5pt}{3pt}
\multirow{2}{*}{$r$}
& \multirow{2}{*}{Method}
& \multicolumn{5}{c}{General}
& \multicolumn{5}{c}{Mathematical Reasoning} \\
\cmidrule(lr){3-7}
\cmidrule(lr){8-12}
&
& HellaSwag & ARC-Challenge & MMLU & C-Eval & Average
& GSM8K & MATH500 & AIME24 & AIME25 & Average \\
\specialrule{1.2pt}{5pt}{3pt}
\multicolumn{12}{c}{\textbf{DeepSeek-V2-Lite}} \\
\midrule
100\%
& Unpruned
& 77.7 & 46.0 & 55.5 & 57.8 & 59.3
& 40.8 & 16.2 & -- & -- & 28.5 \\
\midrule
\multirow{6}{*}{75\%}
& Random
& 60.1 & 35.5 & 36.3 & 35.7 & 41.9
& 9.9 & 4.8 & -- & -- & 7.4 \\
& Frequency
& 74.3 & 40.5 & 36.2 & 29.6 & 45.2
& 36.5 & \textbf{13.2} & -- & -- & 24.9 \\
& EASY-EP
& 65.5 & 37.5 & 35.5 & 35.1 & 43.4
& 25.7 & 6.4 & -- & -- & 16.1 \\
& REAP
& \underline{74.5} & \underline{42.4} & \textbf{45.7} &
\underline{41.5} & \underline{51.0}
& 34.7 & 9.0 & -- & -- & 21.9 \\
& PEEP (Ours)
& \textbf{76.1} & \textbf{43.3} & \underline{45.2} & 39.5 &
\underline{51.0}
& \underline{39.9} & \textbf{13.2} & -- & -- & \underline{26.5} \\
& DCEP (Ours)
& \textbf{76.1} & 41.8 & 44.3 & \textbf{42.7} & \textbf{51.2}
& \textbf{40.8} & \underline{12.6} & -- & -- & \textbf{26.7} \\
\midrule
\multirow{6}{*}{50\%}
& Random
& 37.8 & 24.1 & 24.8 & 23.6 & 27.6
& 2.2 & 1.3 & -- & -- & 1.8 \\
& Frequency
& 46.7 & 25.3 & 24.0 & 22.8 & 29.7
& 20.6 & \underline{6.0} & -- & -- & 13.3 \\
& EASY-EP
& 58.4 & \underline{34.6} & 25.0 & 25.0 & 35.8
& 6.6 & 2.8 & -- & -- & 4.7 \\
& REAP
& \textbf{67.3} & 34.1 & 25.0 & 24.2 & \underline{37.7}
& 11.2 & 3.0 & -- & -- & 7.1 \\
& PEEP (Ours)
& 61.7 & \textbf{36.8} & \textbf{28.1} & \textbf{26.7} &
\textbf{38.3}
& \textbf{23.4} & \textbf{6.2} & -- & -- & \textbf{14.8} \\
& DCEP (Ours)
& \underline{62.4} & 34.1 & \underline{26.8} & \underline{25.6} &
37.2
& \underline{21.8} & 5.8 & -- & -- & \underline{13.8} \\
\specialrule{1.2pt}{5pt}{3pt}
\multicolumn{12}{c}{\textbf{Qwen3-30B-A3B}} \\
\midrule
100\%
& Unpruned
& 79.6 & 63.0 & 80.2 & 86.2 & 77.3
& 86.4 & 90.0 & 73.3 & 70.0 & 79.9 \\
\midrule
\multirow{6}{*}{75\%}
& Random
& 72.2 & 57.0 & 71.6 & \textbf{73.4} & 68.6
& 64.5 & 79.8 & 26.7 & 24.7 & 48.9 \\
& Frequency
& 79.6 & \textbf{62.5} & 74.5 & 68.3 & 71.2
& \underline{88.3} & 83.0 & 63.3 & \underline{60.0} & 73.7 \\
& EASY-EP
& 77.2 & 61.3 & 61.3 & 61.0 & 65.2
& 71.2 & \underline{86.6} & 60.0 & \underline{60.0} & 69.5 \\
& REAP
& 79.4 & 60.7 & 75.1 & \underline{70.0} & 71.3
& 86.9 & \textbf{87.0} & 66.7 & \underline{60.0} & 75.1 \\
& PEEP (Ours)
& \textbf{80.5} & 60.6 & \underline{76.0} & 68.3 &
\underline{71.4}
& \textbf{88.9} & 85.6 & \textbf{76.7} & \textbf{63.3} &
\textbf{78.6} \\
& DCEP (Ours)
& \underline{80.2} & \underline{62.1} & \textbf{76.3} & 68.2 &
\textbf{71.7}
& \textbf{88.9} & \underline{86.6} & \underline{70.0} &
\underline{60.0} & \underline{76.4} \\
\midrule
\multirow{6}{*}{50\%}
& Random
& 56.2 & 38.9 & 51.5 & \textbf{46.3} & 48.2
& 30.1 & 13.6 & 0.0 & 0.0 & 10.9 \\
& Frequency
& 42.7 & 27.1 & 24.6 & 23.0 & 29.4
& 87.8 & 27.4 & \underline{53.3} & 43.3 & 53.0 \\
& EASY-EP
& 70.9 & \textbf{52.0} & 49.3 & 35.7 & 52.0
& 39.1 & 19.2 & \textbf{60.0} & \underline{50.0} & 42.1 \\
& REAP
& 72.6 & \underline{47.7} & 58.8 & 44.6 & 55.9
& 84.4 & \underline{48.0} & 50.0 & 46.7 & 57.3 \\
& PEEP (Ours)
& \textbf{76.1} & 46.6 & \underline{59.7} & 44.4 &
\textbf{56.7}
& \textbf{89.7} & 40.6 & 50.0 & \underline{50.0} &
\underline{57.6} \\
& DCEP (Ours)
& \underline{75.9} & 45.5 & \textbf{60.0} & \underline{44.9} &
\underline{56.6}
& \underline{88.9} & \textbf{51.6} & 50.0 & \textbf{53.3} &
\textbf{60.9} \\
\specialrule{1.2pt}{5pt}{3pt}
\end{tabular}}}
\label{tab:main_results}
\end{table*}

\begin{table*}[htbp]
\centering
\caption{Ablation study of router-derived weights on DeepSeek-V2-Lite and Qwen3-30B-A3B. Columns $freq_i^{(\ell)}$, $\hat{g}_i^{(\ell)}$, and $\bar{g}_i^{(\ell)}$ indicate whether activation frequency, mean conditional routing score, and mean routing score are included in the multiplicative weight construction. A $\checkmark$ indicates that the corresponding router signal is included, whereas a $\times$ indicates that it is excluded. For each model, performance is averaged over all evaluated tasks. The best and second-best results in each column are highlighted in bold and underlined, respectively.}
{
\resizebox{0.6\textwidth}{!}{
\begin{tabular}{ccc cccc cccc}
\toprule
\multicolumn{3}{c}{Router Signals} & \multicolumn{4}{c}{DeepSeek-V2-Lite} & \multicolumn{4}{c}{Qwen3-30B-A3B} \\
\cmidrule(lr){1-3} \cmidrule(lr){4-7} \cmidrule(lr){8-11}
\multirow{2}{*}{$freq_i^{(\ell)}$} & \multirow{2}{*}{$\hat{g}_i^{(\ell)}$} & \multirow{2}{*}{$\bar{g}_i^{(\ell)}$}
& \multicolumn{2}{c}{75\%} & \multicolumn{2}{c}{50\%} & \multicolumn{2}{c}{75\%} & \multicolumn{2}{c}{50\%} \\
\cmidrule(lr){4-5} \cmidrule(lr){6-7} \cmidrule(lr){8-9} \cmidrule(lr){10-11}
& & & PEEP & DCEP & PEEP & DCEP & PEEP & DCEP & PEEP & DCEP \\
\midrule
$\times$ & $\times$ & $\times$ & \textbf{43.6} & 42.6 & 29.8 & 29.4 & 70.0 & 70.0 & 46.1 & 51.0 \\
$\checkmark$ & $\checkmark$ & $\checkmark$ & 42.2 & \textbf{43.9} & 30.0 & \underline{30.4} & \textbf{75.5} & \textbf{75.0} & 57.7 & \underline{58.3} \\
$\checkmark$ & $\times$ & $\times$ & 42.9 & \underline{43.1} & 30.5 & 29.4 & \underline{75.0} & \underline{74.0} & 57.1 & \textbf{58.8} \\
$\times$ & $\checkmark$ & $\times$ & \underline{43.2} & 43.0 & 24.9 & 29.4 & 70.2 & 72.4 & 55.4 & 54.8 \\
$\times$ & $\times$ & $\checkmark$ & 42.9 & 42.4 & \textbf{31.1} & \textbf{30.7} & 73.7 & 73.8 & 57.0 & \textbf{58.8} \\
$\checkmark$ & $\checkmark$ & $\times$ & 42.0 & 42.0 & \underline{30.6} & 30.1 & 71.3 & 71.0 & \underline{57.8} & 58.0 \\
$\checkmark$ & $\times$ & $\checkmark$ & \underline{43.2} & 42.6 & 29.2 & 28.8 & 71.3 & 71.8 & 57.4 & 56.3 \\
$\times$ & $\checkmark$ & $\checkmark$ & 42.4 & 42.6 & 25.7 & 25.1 & 71.9 & 72.4 & \textbf{58.3} & 58.0 \\
\bottomrule
\end{tabular}}}
\label{tab:ab_weight}
\end{table*}

\begin{figure*}[ht]
\centering
\includegraphics[width=0.9\textwidth]{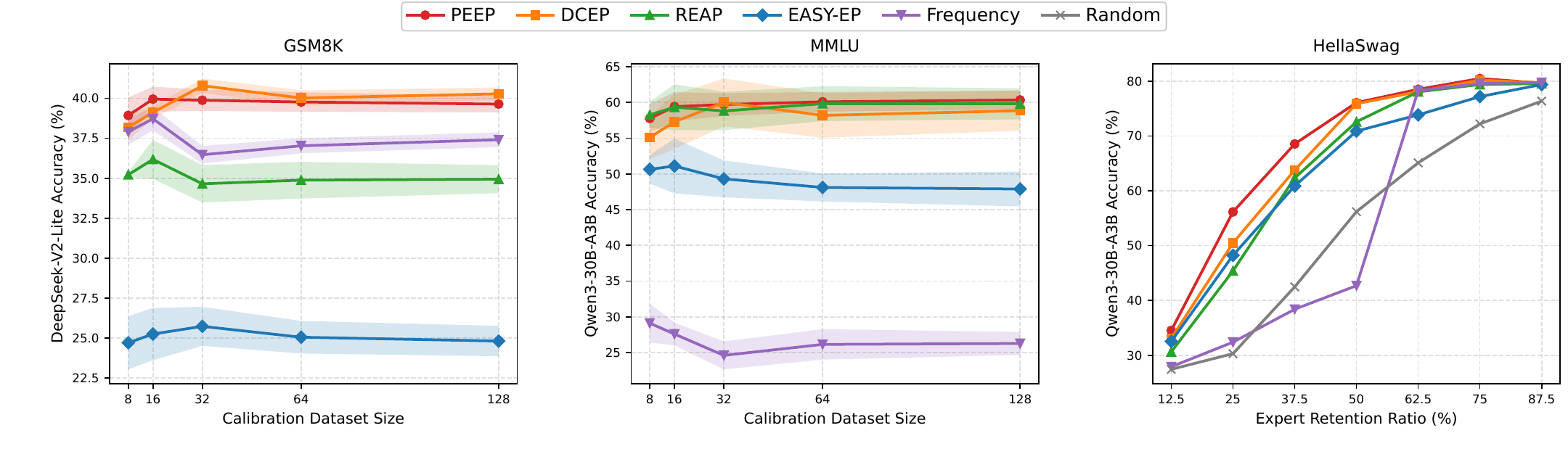}
\caption{\textit{Left}: Performance comparison among different pruning methods with different calibration dataset sizes on DeepSeek-V2-Lite model and GSM8K task at expert retention ratio of $r{=}75\%$. \textit{Middle}: Performance comparison among different pruning methods with different calibration dataset sizes on Qwen3-30B-A3B model and MMLU task at expert retention ratio of $r{=}50\%$. \textit{Right}: Performance comparison among different pruning methods with different expert retention ratios on Qwen3-30B-A3B model and HellaSwag task.}
\label{fig:abs}
\vspace{-1em}
\end{figure*}

\subsection{Experimental Results}
\paragraph{Main Results}
Table~\ref{tab:main_results} presents a performance comparison of different expert pruning methods on DeepSeek-V2-Lite and Qwen3-30B-A3B across general and mathematical reasoning benchmarks at expert retention ratios of $r{=}50\%$ and $r{=}75\%$. 

Overall, our structure-preserving expert pruning methods consistently achieve strong performance across both general and mathematical reasoning tasks. Notably, among all eight combinations of model, expert retention ratio, and task category, the best average performance is always achieved by either PEEP or DCEP. The advantage is particularly pronounced on mathematical reasoning tasks with Qwen3-30B-A3B. At expert retention ratio of $r{=}75\%$, PEEP achieves a mathematical reasoning average performance of 78.6\%, outperforming the SOTA REAP baseline by 3.5 percentage points and remaining only 1.3 percentage points below the unpruned model. At more aggressive expert retention ratio of $r{=}50\%$, DCEP retains a mathematical reasoning average performance of 60.9\% on Qwen3-30B-A3B, exceeding REAP by 3.6 percentage points.

\vspace{-0.5em}
\paragraph{Ablation Study of Router-Derived Weights}
Table~\ref{tab:ab_weight} presents a performance comparison of PEEP and DCEP under different variants of router-derived weights. The complete ablation results are provided in Tables~\ref{tab:peep_weight_ablation} and~\ref{tab:dcep_weight_ablation} in Appendix~\ref{app:router_weight_ablation}. The ablation results show that activation frequency and mean routing score provide effective router-derived weights as priors of expert importance. We attribute this to the complementary roles of global structural modeling and local importance estimation: PEEP and DCEP capture the global structure of functional subspace and distributional geometry among experts, whereas activation frequency and mean routing score provide stable estimates of the local importance of individual basis directions and distributional modes. Although multiplying all router signals yields the best overall performance, we adopt activation frequency as the default weighting signal in our experiments, considering both effectiveness and simplicity.

\vspace{-0.5em}
\paragraph{Effect of Calibration Dataset Sizes}
Figure~\ref{fig:abs} illustrates the effect of calibration dataset sizes, with the complete results provided in Figures~\ref{fig:ab_size_75} and \ref{fig:ab_size_50} in Appendix~\ref{app:calibration_size}. respectively. Increasing the calibration set from a very small size generally improves the reliability of pruning and reduces the variation. Using 32 calibration samples already achieves stable results, while increasing the size to 64 or 128 yields no consistent benefit. This suggests that the structural statistics exploited by our methods can be estimated reliably from only a small calibration set, making 32 samples a favorable trade-off between calibration cost and pruning quality.

\vspace{-0.5em}
\paragraph{Effect of Expert Retention Ratios}
Figure~\ref{fig:abs} illustrates the effect of expert retention ratios, with the complete results provided in Figure~\ref{fig:ab_ratio} in Appendix~\ref{app:retention_ratio}. As expected, performance generally improves as more experts are retained, while aggressive pruning leads to a marked degradation. Across a wide range of expert retention ratios, our methods are consistently competitive, showing that preserving the structure of expert outputs is particularly beneficial.

\vspace{-0.5em}
\section{Conclusion}
We study expert pruning for MoE LLMs in a structure-preserving framework. We instantiate the structure-preserving framework with PEEP and DCEP, which maximize projection energy in the functional subspace and minimize coverage error over the distributional modes, respectively. On DeepSeek-V2-Lite and Qwen3-30B-A3B, our methods consistently achieve the best average performance on general and mathematical reasoning benchmarks at 75\% and 50\% expert retention ratios.



\subsection*{AI use statement}
In this work, we used generative AI tools for implementing method, assisting with translation. We have not used generative AI tools for generating synthetic data sets, helping develop theoretical models or conceptual frameworks, formulating mathematical claims, providing critical ingredients for proving mathematical claims, assisting in the writing of proofs, proposing or refining hypotheses, designing or providing feedback on research  methodology or experiments, cleaning and reformatting dataset, interpreting results. Supporting qualitative and thematic data analysis is not applicable to this work. Additionally, we used generative AI tools for creating or modifying scientific figures or images, vibe coding. We have reviewed all AI-assisted work. We reviewed and verified all AI-assisted content to ensure the correctness. We take responsibility for the final content of this work, including text, claims or artifacts produced with the aid of generative AI.

\subsection*{Reproducibility statement}
We show our complete expert pruning pipeline in Figure~\ref{fig:method} and describe the detailed algorithms for PEEP and DCEP in Algorithms~\ref{alg:peep} and~\ref{alg:dcep}. We provide the detailed experimental setup in Section~\ref{sec:experiments} and Appendix~\ref{app:experimental_setup}. We also provide the source code for reproducing the experimental results in supplementary materials.

\bibliography{iclr2027_conference}
\bibliographystyle{iclr2027_conference}

\clearpage
\appendix
\section{Detailed Experimental Setup}
\label{app:experimental_setup}
\vspace{-0.5em}
\paragraph{Models}
We conduct experiments on DeepSeek-V2-Lite and Qwen3-30B-A3B. For each model, we first perform a forward pass over a calibration dataset to probe the MoE layers and collect the mean and standard deviation of each expert's output. We then apply our proposed PEEP and DCEP methods to prune experts under two retention ratios, i.e., retaining 75\% and 50\% of the experts in every MoE layer. Finally, we evaluate the resulting pruned models.

\vspace{-0.5em}
\paragraph{Datasets}
We evaluate the pruned models on two categories of downstream tasks. The first category comprises general benchmarks, including HellaSwag~\citep{HellaSwag}, ARC-Challenge~\citep{ARC}, MMLU~\citep{MMLU}, and C-Eval~\citep{CEval}. These benchmarks evaluate commonsense reasoning, scientific question answering, broad-domain knowledge, and Chinese-language understanding, respectively, providing diverse coverage of general model capabilities. For these benchmarks, we use a shared calibration dataset consisting of 32 samples drawn from C4~\citep{C4}. The second category comprises mathematical reasoning benchmarks, including GSM8K~\citep{GSM8K}, MATH500~\citep{MATH500}, AIME24~\citep{AIME2024}, and AIME25~\citep{AIME2025}. These benchmarks span a broad range of difficulty, from elementary multi-step reasoning to competition-level mathematical problem solving. For GSM8K and MATH500, we use 32 samples drawn from the training split of GSM8K as the calibration dataset. For AIME24 and AIME25, we use the 30 problems from AIME23 as the calibration dataset. All evaluations are conducted on four NVIDIA A100 40GB GPUs using the vLLM backend~\citep{vLLM} and the LM-Evaluation-Harness framework~\citep{LM_Evaluation_Harness}. 

The selection of these evaluation benchmarks enables us to assess not only the overall robustness of the pruning methods across domains and languages, but also their ability to preserve complex reasoning capabilities that may depend more strongly on coordination among experts.

\vspace{-0.5em}
\paragraph{Baselines}
We compare our methods against random pruning, frequency-based pruning, EASY-EP~\citep{EASY_EP}, and REAP~\citep{REAP}. Random pruning and frequency-based pruning serve as simple and intuitive baselines, where experts are selected randomly or according to their activation frequencies. EASY-EP and REAP represent recent SOTA methods.

\vspace{-0.5em}
\section{Full Ablation Results for Router-Derived Weights}
\label{app:router_weight_ablation}
Tables~\ref{tab:peep_weight_ablation} and~\ref{tab:dcep_weight_ablation} present the complete performance comparison of PEEP and DCEP under different variants of router-derived weights. Overall, activation frequency and mean routing score provide effective priors of expert importance across different models and pruning settings. These results are consistent with Table~\ref{tab:ab_weight}. Although multiplying all router signals yields the best overall performance, we adopt activation frequency as the default weighting signal in our experiments, considering both effectiveness and simplicity.

\begin{table*}[htbp]
\centering
\caption{
Ablation study of router-derived weights for PEEP on DeepSeek-V2-Lite and Qwen3-30B-A3B. Columns $freq_i^{(\ell)}$, $\hat{g}_i^{(\ell)}$, and $\bar{g}_i^{(\ell)}$ indicate whether activation frequency, mean conditional routing score, and mean routing score are included in the multiplicative weight construction. ``$\checkmark$'' indicates that the corresponding router signal is included in the weights, whereas ``$\times$'' indicates that it is not included. The best and second-best results among pruned methods under each model and expert retention ratio are highlighted in bold and underlined, respectively. ``--'' indicates that the model fails to complete the corresponding task.
}
{
\resizebox{1.0\textwidth}{!}{
\begin{tabular}{lccccccccccccc}
\specialrule{1.2pt}{5pt}{3pt}
\multirow{2}{*}{$r$}
& \multicolumn{3}{c}{Router Signals}
& \multicolumn{5}{c}{General}
& \multicolumn{5}{c}{Mathematical Reasoning} \\
\cmidrule(lr){2-4}
\cmidrule(lr){5-9}
\cmidrule(lr){10-14}
&
$freq_i^{(\ell)}$ & $\hat{g}_i^{(\ell)}$ & $\bar{g}_i^{(\ell)}$
& HellaSwag & ARC-Challenge & MMLU & C-Eval & Average
& GSM8K & MATH500 & AIME24 & AIME25 & Average \\
\specialrule{1.2pt}{5pt}{3pt}
\multicolumn{14}{c}{\textbf{DeepSeek-V2-Lite}} \\
\midrule
\multirow{8}{*}{75\%}
& $\times$ & $\times$ & $\times$
& 73.9 & 41.6 & \textbf{47.2} & 46.3 & \underline{52.2}
& 37.0 & \textbf{15.8} & -- & -- & 26.4 \\
& $\checkmark$ & $\checkmark$ & $\checkmark$
& 75.2 & 41.0 & 45.2 & 41.9 & 50.8
& 36.8 & 13.0 & -- & -- & 24.9 \\
& $\checkmark$ & $\times$ & $\times$
& \textbf{76.1} & 43.3 & 45.2 & 39.5 & 51.0
& 39.9 & 13.2 & -- & -- & \underline{26.5} \\
& $\times$ & $\checkmark$ & $\times$
& \underline{75.3} & \textbf{44.8} & \underline{45.5} & \textbf{46.6} & \textbf{53.0}
& 36.9 & 10.2 & -- & -- & 23.6 \\
& $\times$ & $\times$ & $\checkmark$
& 74.9 & \underline{41.8} & 45.4 & 42.7 & 51.2
& 39.1 & 13.2 & -- & -- & 26.2 \\
& $\checkmark$ & $\checkmark$ & $\times$
& 74.9 & 40.9 & 44.0 & 42.9 & 50.7
& 37.6 & 11.6 & -- & -- & 24.6 \\
& $\checkmark$ & $\times$ & $\checkmark$
& \underline{75.3} & 41.1 & 44.9 & 44.0 & 51.3
& \underline{39.8} & \underline{14.2} & -- & -- & \textbf{27.0} \\
& $\times$ & $\checkmark$ & $\checkmark$
& 74.8 & 40.3 & 44.4 & 43.8 & 50.8
& 39.3 & 11.6 & -- & -- & 25.5 \\
\midrule
\multirow{8}{*}{50\%}
& $\times$ &  $\times$ & $\times$
& 62.7 & 33.3 & 28.9 & \textbf{30.9} & 38.9
& 19.0 & 4.0 & -- & -- & 11.5 \\
& $\checkmark$ & $\checkmark$ & $\checkmark$
& 64.0 & 33.7 & 26.3 & 28.0 & 38.0
& \textbf{23.4} & 4.8 & -- & -- & \underline{14.1} \\
& $\checkmark$ & $\times$ & $\times$
& 61.7 & \textbf{36.8} & 28.1 & 26.7 & 38.3
& \textbf{23.4} & \textbf{6.2} & -- & -- & \textbf{14.8} \\
& $\times$ & $\checkmark$ & $\times$
& 51.7 & 25.4 & 24.0 & 25.0 & 31.5
& 17.6 & 5.6 & -- & -- & 11.6 \\
& $\times$ & $\times$ & $\checkmark$
& \underline{65.9} & 33.0 & \textbf{31.2} & \underline{28.7} & \textbf{39.7}
& \underline{21.8} & \underline{5.8} & -- & -- & 13.8 \\
& $\checkmark$ & $\checkmark$ & $\times$
& \textbf{66.3} & \underline{35.9} & \underline{28.4} & 26.6 & \underline{39.3}
& 21.1 & 5.4 & -- & -- & 13.2 \\
& $\checkmark$ & $\times$ & $\checkmark$
& 63.7 & 35.4 & 25.9 & 23.6 & 37.1
& 20.8 & \underline{5.8} & -- & -- & 13.3 \\
& $\times$ & $\checkmark$ & $\checkmark$
& 51.5 & 26.0 & 24.2 & 25.0 & 31.7
& \underline{21.8} & 5.6 & -- & -- & 13.7 \\
\specialrule{1.2pt}{5pt}{3pt}
\multicolumn{14}{c}{\textbf{Qwen3-30B-A3B}} \\
\midrule
\multirow{8}{*}{75\%}
& $\times$ & $\times$ & $\times$
& 67.1 & 57.9 & 68.6 & \textbf{71.3} & 66.2
& 86.3 & 82.6 & 63.3 & 63.3 & 73.9 \\
& $\checkmark$ & $\checkmark$ & $\checkmark$
& 79.5 & \textbf{61.1} & 75.3 & 67.8 & 70.9
& \underline{89.1} & \underline{88.0} & \underline{73.3} & \textbf{70.0} & \textbf{80.1} \\
& $\checkmark$ & $\times$ & $\times$
& \textbf{80.5} & 60.6 & \textbf{76.0} & 68.3 & \textbf{71.4}
& 88.9 & 85.6 & \textbf{76.7} & 63.3 & \underline{78.6} \\
& $\times$ & $\checkmark$ & $\times$
& 79.5 & 58.1 & 71.7 & \underline{68.8} & 69.5
& 88.1 & 85.8 & 60.0 & 50.0 & 71.0 \\
& $\times$ & $\times$ & $\checkmark$
& 79.4 & 60.6 & 74.8 & 66.3 & 70.3
& \textbf{89.7} & 85.6 & 66.7 & \underline{66.7} & 77.2 \\
& $\checkmark$ & $\checkmark$ & $\times$
& 79.3 & 61.0 & 74.8 & 66.6 & 70.4
& 83.6 & \textbf{88.4} & 60.0 & 56.7 & 72.2 \\
& $\checkmark$ & $\times$ & $\checkmark$
& \underline{79.7} & \textbf{61.1} & \underline{75.2} & 68.3 & \underline{71.1}
& 80.4 & 85.8 & 60.0 & 60.0 & 71.5 \\
& $\times$ & $\checkmark$ & $\checkmark$
& 79.4 & 60.8 & 75.1 & 67.4 & 70.7
& 80.7 & 88.2 & 66.7 & 56.7 & 73.1 \\
\midrule
\multirow{8}{*}{50\%}
& $\times$ & $\times$ & $\times$
& 40.6 & 40.0 & 45.2 & \textbf{50.7} & 44.1
& 80.7 & 45.2 & 40.0 & 26.7 & 48.1 \\
& $\checkmark$ & $\checkmark$ & $\checkmark$
& \textbf{76.7} & \textbf{47.7} & 59.6 & 45.8 & \textbf{57.5}
& 88.2 & 33.4 & \textbf{63.3} & 46.7 & 57.9 \\
& $\checkmark$ & $\times$ & $\times$
& \underline{76.1} & \underline{46.6} & 59.7 & 44.4 & \underline{56.7}
& \textbf{89.7} & 40.6 & 50.0 & \textbf{50.0} & 57.6 \\
& $\times$ & $\checkmark$ & $\times$
& 51.6 & 42.6 & 51.0 & \underline{46.4} & 47.9
& 86.4 & \textbf{55.0} & \underline{60.0} & \textbf{50.0} & \textbf{62.8} \\
& $\times$ & $\times$ & $\checkmark$
& 75.5 & 43.6 & 59.9 & 44.7 & 55.9
& \underline{89.6} & 39.2 & 56.7 & 46.7 & 58.0 \\
& $\checkmark$ & $\checkmark$ & $\times$
& 75.4 & 43.8 & \textbf{60.1} & 44.9 & 56.0
& 86.4 & 45.0 & 56.7 & \textbf{50.0} & 59.5 \\
& $\checkmark$ & $\times$ & $\checkmark$
& 76.6 & 46.9 & 59.5 & 43.3 & 56.6
& 87.3 & 35.8 & \textbf{63.3} & 46.7 & 58.3 \\
& $\times$ & $\checkmark$ & $\checkmark$
& 75.3 & 44.5 & \underline{59.8} & 45.5 & 56.3
& 85.4 & 46.2 & \textbf{63.3} & 46.7 & \underline{60.4} \\
\specialrule{1.2pt}{5pt}{3pt}
\end{tabular}}}
\label{tab:peep_weight_ablation}
\end{table*}

\begin{table*}[htbp]
\centering
\caption{
Ablation study of router-derived weights for DCEP on DeepSeek-V2-Lite and Qwen3-30B-A3B. Columns $freq_i^{(\ell)}$, $\hat{g}_i^{(\ell)}$, and $\bar{g}_i^{(\ell)}$ indicate whether activation frequency, mean conditional routing score, and mean routing score are included in the multiplicative weight construction. ``$\checkmark$'' indicates that the corresponding router signal is included in the weights, whereas ``$\times$'' indicates that it is not included. The best and second-best results among pruned methods under each model and expert retention ratio are highlighted in bold and underlined, respectively. ``--'' indicates that the model fails to complete the corresponding task.
}
{
\resizebox{1.0\textwidth}{!}{
\begin{tabular}{lccccccccccccc}
\specialrule{1.2pt}{5pt}{3pt}
\multirow{2}{*}{$r$}
& \multicolumn{3}{c}{Router Signals}
& \multicolumn{5}{c}{General}
& \multicolumn{5}{c}{Mathematical Reasoning} \\
\cmidrule(lr){2-4}
\cmidrule(lr){5-9}
\cmidrule(lr){10-14}
&
$freq_i^{(\ell)}$ & $\hat{g}_i^{(\ell)}$ & $\bar{g}_i^{(\ell)}$
& HellaSwag & ARC-Challenge & MMLU & C-Eval & Average
& GSM8K & MATH500 & AIME24 & AIME25 & Average \\
\specialrule{1.2pt}{5pt}{3pt}
\multicolumn{14}{c}{\textbf{DeepSeek-V2-Lite}} \\
\midrule
\multirow{8}{*}{75\%}
& $\times$ & $\times$ & $\times$
& 75.4 & 40.4 & 44.6 & 43.7 & 51.0
& 38.5 & 12.8 & -- & -- & 25.7 \\
& $\checkmark$ & $\checkmark$ & $\checkmark$
& \underline{75.5} & 43.1 & \textbf{47.1} & \textbf{47.8} & \textbf{53.4}
& 36.8 & 13.0 & -- & -- & 24.9 \\
& $\checkmark$ & $\times$ & $\times$
& \textbf{76.1} & 41.8 & 44.3 & 42.7 & 51.2
& \textbf{40.8} & 12.6 & -- & -- & \textbf{26.7} \\
& $\times$ & $\checkmark$ & $\times$
& 75.0 & \textbf{44.1} & \underline{46.2} & \underline{44.7} & \underline{52.5}
& 37.4 & 10.6 & -- & -- & 24.0 \\
& $\times$ & $\times$ & $\checkmark$
& 75.0 & 40.4 & 45.3 & 42.6 & 50.8
& \underline{38.4} & 13.0 & -- & -- & 25.7 \\
& $\checkmark$ & $\checkmark$ & $\times$
& 74.9 & 40.2 & 44.3 & 43.9 & 50.8
& 36.7 & 11.8 & -- & -- & 24.2 \\
& $\checkmark$ & $\times$ & $\checkmark$
& \underline{75.5} & 40.7 & 44.3 & 43.2 & 50.9
& 38.1 & \textbf{13.6} & -- & -- & \underline{25.8} \\
& $\times$ & $\checkmark$ & $\checkmark$
& 75.2 & \underline{42.0} & 44.8 & \textbf{45.2} & 51.8
& 38.1 & 10.6 & -- & -- & 24.3 \\
\midrule
\multirow{8}{*}{50\%}
& $\times$ & $\times$ & $\times$
& 64.1 & 32.3 & 26.3 & 27.0 & 37.4
& 21.4 & 5.2 & -- & -- & 13.3 \\
& $\checkmark$ & $\checkmark$ & $\checkmark$
& 63.8 & \textbf{35.5} & \underline{29.1} & \underline{27.9} & \textbf{39.1}
& 21.5 & 4.4 & -- & -- & 12.9 \\
& $\checkmark$ & $\times$ & $\times$
& 62.4 & 34.1 & 26.8 & 25.6 & 37.2
& 21.8 & \textbf{5.8} & -- & -- & 13.8 \\
& $\times$ & $\checkmark$ & $\times$
& \underline{65.0} & 32.6 & 28.2 & \textbf{28.2} & 38.5
& 16.8 & \textbf{5.8} & -- & -- & 11.3 \\
& $\times$ & $\times$ & $\checkmark$
& 64.5 & 33.2 & \textbf{30.5} & 27.2 & \underline{38.8}
& \textbf{24.4} & 4.4 & -- & -- & \textbf{14.4} \\
& $\checkmark$ & $\checkmark$ & $\times$
& \textbf{66.2} & \underline{35.0} & 27.0 & 25.5 & 38.4
& 21.0 & \underline{5.6} & -- & -- & 13.3 \\
& $\checkmark$ & $\times$ & $\checkmark$
& 64.1 & 32.9 & 24.6 & 22.9 & 36.1
& \underline{23.5} & 5.0 & -- & -- & \underline{14.3} \\
& $\times$ & $\checkmark$ & $\checkmark$
& 51.8 & 26.2 & 23.6 & 24.4 & 31.5
& 20.6 & 4.2 & -- & -- & 12.4 \\
\specialrule{1.2pt}{5pt}{3pt}
\multicolumn{14}{c}{\textbf{Qwen3-30B-A3B}} \\
\midrule
\multirow{8}{*}{75\%}
& $\times$ & $\times$ & $\times$
& 69.1 & 56.5 & 69.0 & 68.1 & 65.7
& 88.0 & 82.6 & \textbf{76.7} & 50.0 & 74.3 \\
& $\checkmark$ & $\checkmark$ & $\checkmark$
& 79.6 & 61.9 & 75.5 & \textbf{70.4} & \textbf{71.8}
& \textbf{89.2} & 87.0 & \underline{73.3} & \underline{63.3} & \textbf{78.2} \\
& $\checkmark$ & $\times$ & $\times$
& \textbf{80.2} & \underline{62.1} & \textbf{76.3} & 68.2 & \underline{71.7}
& \underline{88.9} & 86.6 & 70.0 & 60.0 & 76.4 \\
& $\times$ & $\checkmark$ & $\times$
& 76.4 & 57.7 & 72.6 & \underline{69.1} & 69.0
& 87.9 & 85.8 & 63.3 & \textbf{66.7} & 75.9 \\
& $\times$ & $\times$ & $\checkmark$
& 79.4 & 60.8 & 75.1 & 66.5 & 70.4
& 88.4 & 86.4 & 66.7 & \textbf{66.7} & \underline{77.0} \\
& $\checkmark$ & $\checkmark$ & $\times$
& 79.4 & 60.2 & 75.1 & 65.4 & 70.0
& 79.8 & \underline{88.0} & 60.0 & 60.0 & 72.0 \\
& $\checkmark$ & $\times$ & $\checkmark$
& 79.6 & \textbf{63.1} & 75.5 & 66.9 & 71.2
& 76.6 & 86.0 & 66.7 & 60.0 & 72.3 \\
& $\times$ & $\checkmark$ & $\checkmark$
& 79.4 & 59.8 & 75.3 & 65.9 & 70.1
& 81.4 & \textbf{90.4} & 70.0 & 56.7 & 74.6 \\
\midrule
\multirow{8}{*}{50\%}
& $\times$ & $\times$ & $\times$
& 45.8 & 33.5 & 47.0 & \textbf{47.3} & 43.4
& 86.9 & \underline{50.6} & 43.3 & 53.3 & 58.5 \\
& $\checkmark$ & $\checkmark$ & $\checkmark$
& 76.1 & \textbf{48.8} & 59.8 & \textbf{45.5} & \textbf{57.5}
& 87.3 & 32.2 & \textbf{73.3} & 43.3 & 59.1 \\
& $\checkmark$ & $\times$ & $\times$
& 75.9 & 45.5 & 60.0 & 44.9 & \underline{56.6}
& \textbf{88.9} & \textbf{51.6} & 50.0 & 53.3 & 60.9 \\
& $\times$ & $\checkmark$ & $\times$
& 58.3 & 37.5 & 52.6 & 43.6 & 48.0
& 85.7 & 47.2 & 56.7 & \underline{56.7} & \textbf{61.6} \\
& $\times$ & $\times$ & $\checkmark$
& 76.0 & 44.7 & \underline{60.4} & 45.1 & \underline{56.6}
& \underline{88.1} & 46.0 & 60.0 & 50.0 & \underline{61.0} \\
& $\checkmark$ & $\checkmark$ & $\times$
& 75.9 & 45.3 & \textbf{60.6} & 44.2 & 56.5
& 85.8 & 45.4 & 46.7 & \textbf{60.0} & 59.5 \\
& $\checkmark$ & $\times$ & $\checkmark$
& \textbf{76.4} & \underline{46.1} & 59.1 & 44.2 & 56.4
& 86.4 & 34.6 & \underline{63.3} & 40.0 & 56.1 \\
& $\times$ & $\checkmark$ & $\checkmark$
& 75.3 & 43.0 & 59.6 & 44.7 & 55.6
& 86.4 & 38.0 & 56.7 & \textbf{60.0} & 60.3 \\
\specialrule{1.2pt}{5pt}{3pt}
\end{tabular}}}
\label{tab:dcep_weight_ablation}
\end{table*}

\vspace{-0.5em}
\section{Effect of Calibration Dataset Sizes}
\label{app:calibration_size}
\begin{figure*}[htbp]
\centering
\includegraphics[width=0.9\textwidth]{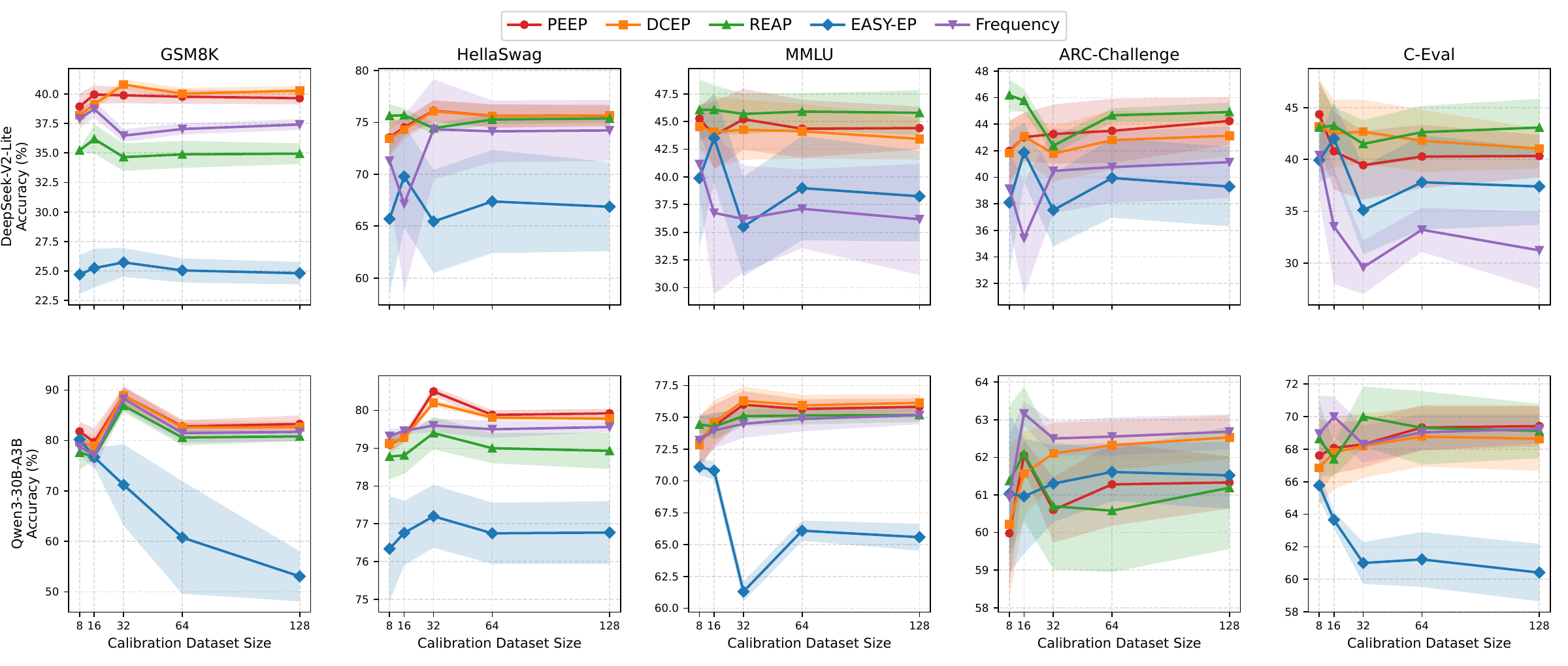}
\caption{Performance comparison among different pruning methods
with different calibration dataset sizes at expert retention ratio of $r{=}75\%$. Results are averaged over five runs with calibration datasets independently sampled using different random seeds. The shaded regions represent $\pm$ one standard deviation. \textit{Top}: DeepSeek-V2-Lite. \textit{Bottom}: Qwen3-30B-A3B. \textit{From left to right}: GSM8K, HellaSwag, MMLU, ARC-Challenge, and C-Eval. }
\label{fig:ab_size_75}
\end{figure*}

\begin{figure*}[htbp]
\centering
\includegraphics[width=0.9\textwidth]{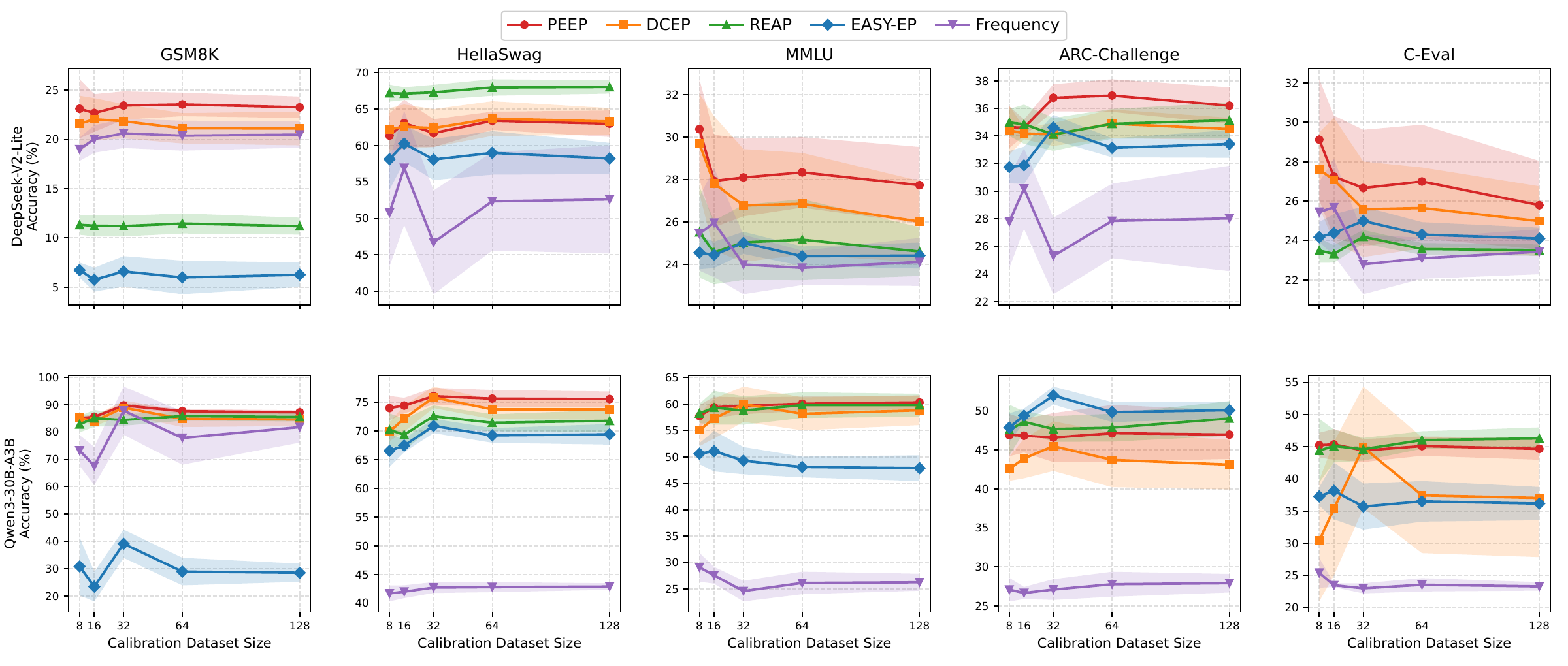}
\caption{Performance comparison among different pruning methods
with different calibration dataset sizes at expert retention ratio of $r{=}50\%$. Results are averaged over five runs with calibration datasets independently sampled using different random seeds. The shaded regions represent $\pm$ one standard deviation. \textit{Top}: DeepSeek-V2-Lite. \textit{Bottom}: Qwen3-30B-A3B. \textit{From left to right}: GSM8K, HellaSwag, MMLU, ARC-Challenge, and C-Eval.} 
\label{fig:ab_size_50}
\end{figure*}

Figures~\ref{fig:ab_size_75} and~\ref{fig:ab_size_50} present the complete performance comparison of different pruning methods across calibration dataset sizes at expert retention ratios of $r{=}75\%$ and $r{=}50\%$. Increasing the calibration dataset from a very small size generally improves the reliability of pruning and reduces the variation. Using 32 calibration samples already achieves stable performance, while further increasing the size to 64 or 128 yields no consistent benefit  These results show that the structural statistics exploited by PEEP and DCEP can be estimated reliably from a small calibration dataset, making 32 samples a favorable trade-off between calibration cost and pruning quality.

\vspace{-0.5em}
\section{Effect of Expert Retention Ratios}
\label{app:retention_ratio}
\begin{figure*}[htbp]
\centering
\includegraphics[width=0.9\textwidth]{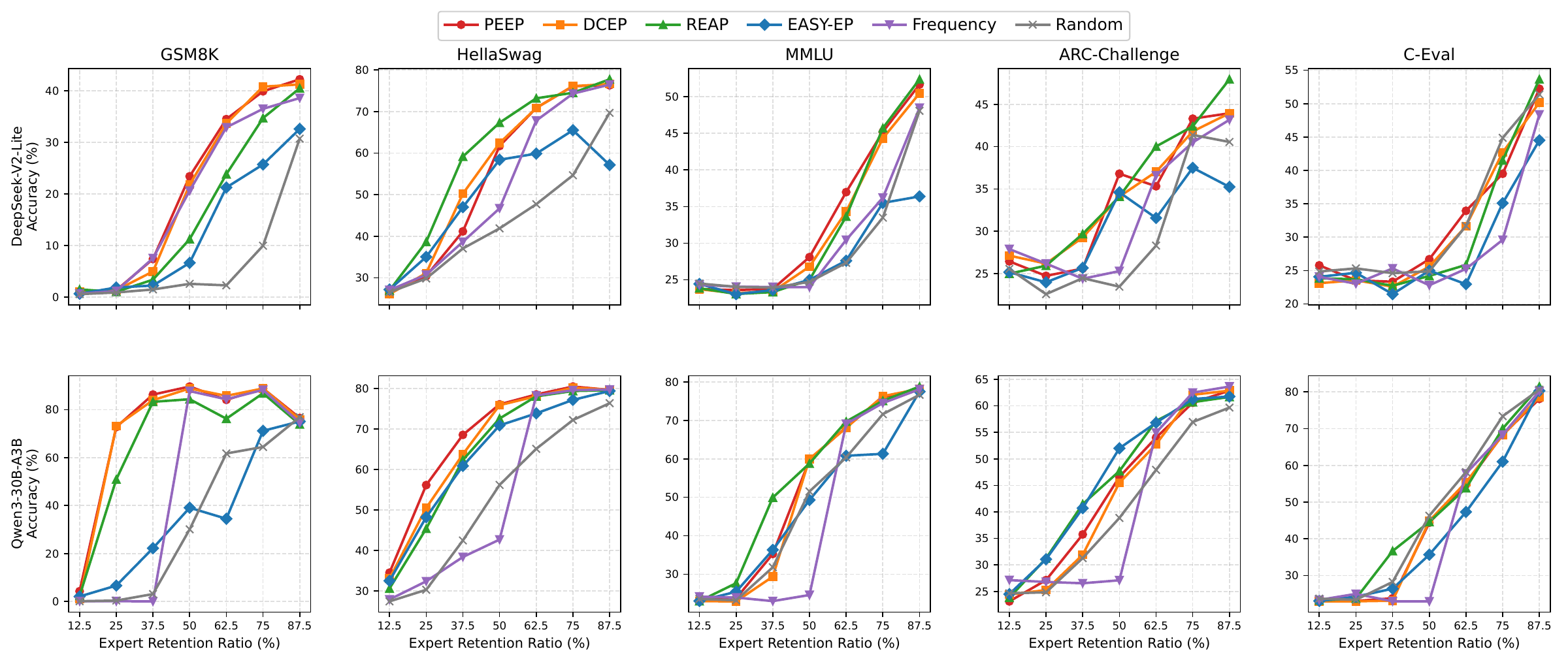}
\caption{Performance comparison among different pruning methods
with different expert retention ratios. \textit{Top}: DeepSeek-V2-Lite. \textit{Bottom}: Qwen3-30B-A3B. \textit{From left to right}: GSM8K, HellaSwag, MMLU, ARC-Challenge, and C-Eval.}
\label{fig:ab_ratio}
\end{figure*}

Figure~\ref{fig:ab_ratio} presents the complete performance comparison of different pruning methods across expert retention ratios. As expected, performance generally improves as more experts are retained, while aggressive pruning leads to a marked degradation. Across a wide range of expert retention ratios, PEEP and DCEP remain consistently competitive with the baselines. These results show that preserving the functional subspace or distributional geometry of expert outputs is beneficial across different pruning budgets.

\section{Sensitivity to the Scaling Factors}
\label{app:scaling_factor}
\begin{figure*}[htbp]
\centering
\includegraphics[width=0.9\textwidth]{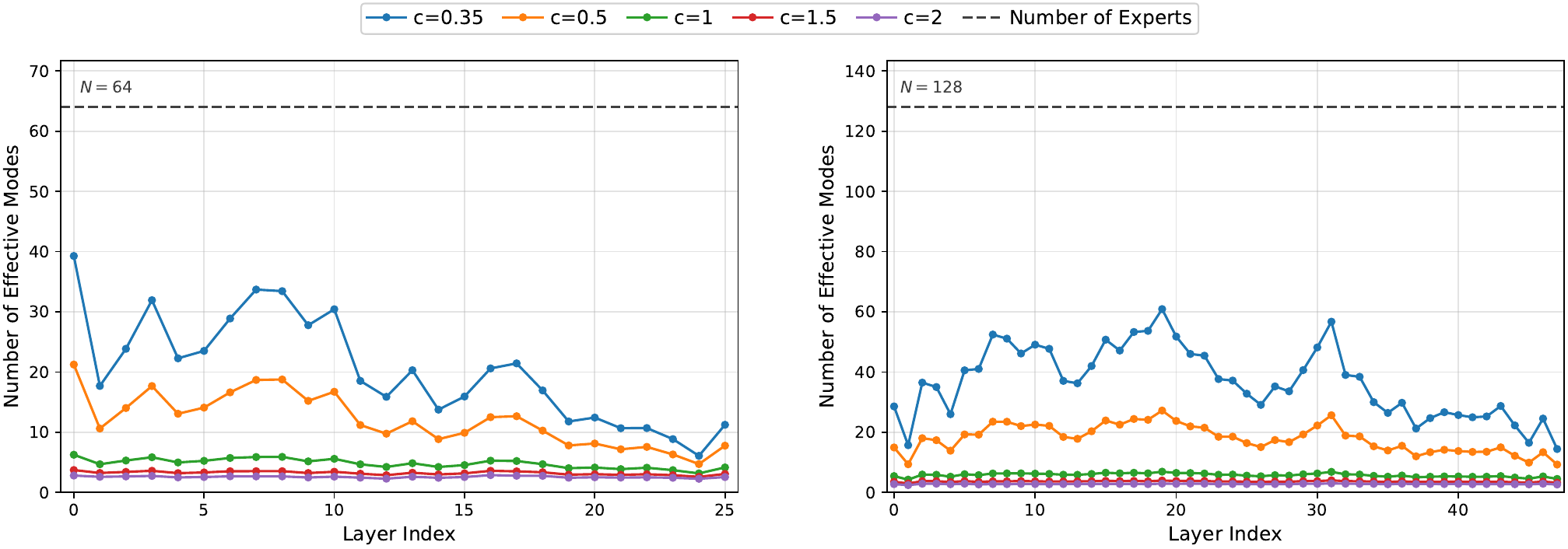}
\caption{Numbers of effective modes of expert output distributions across MoE layers under different values of scaling factor $c$.
\textit{Left:} DeepSeek-V2-Lite with 64 experts per layer.
\textit{Right:} Qwen3-30B-A3B with 128 experts per layer.
Across all MoE layers and different values of scaling factor $c$, the number of effective modes remains well below the nominal number of experts.}
\label{fig:ab_neff}
\end{figure*}

Figure~\ref{fig:ab_neff} presents the numbers of effective distributional modes across MoE layers under different values of the scaling factor $c$. We evaluate $c\in\{0.35,0.5,1,1.5,2\}$ on DeepSeek-V2-Lite and Qwen3-30B-A3B, and use $c{=}0.5$ for the results reported in Figure~\ref{fig:motivation} in the main text. Increasing $c$ reduces the estimated number of effective modes because a larger kernel bandwidth assigns higher similarity to different expert distributions. Although the absolute value varies with $c$, the number of effective modes remains substantially smaller than the total number of experts across all MoE layers and all tested values of $c$. This shows that the observed distributional redundancy is robust to the choice of the scaling factor.

\end{document}